\documentclass{thesis}

\usepackage{latexsym}
\usepackage{enumerate}
\usepackage{booktabs}
\usepackage{amsmath,amsfonts,amssymb,amsthm}

\usepackage[utf8]{inputenc}

\usepackage{microtype}

\usepackage{pifont}     % Used for xmark and cmark
\usepackage{natbib}     % Used to get different cite cmds
\usepackage{graphicx}   % For resizebox

\newcommand{\cmark}{\ding{51}}
\newcommand{\xmark}{\ding{55}}

\newcommand\tf[1]{\textbf{#1}}

\newcommand{\cls}{\texttt{[CLS]}}
\newcommand{\sep}{\texttt{[SEP]}}

\author{Christopher Sciavolino}
\adviser{Danqi Chen}
\title{Towards Universal Dense Retrieval for Open-domain Question Answering}
\abstract{%!TEX root = ../thesis.tex

In open-domain question answering, a model receives a text question as input and searches for the correct answer using a large evidence corpus. The retrieval step is especially difficult as typical evidence corpora have \textit{millions} of documents, each of which may or may not have the correct answer to the question.

Very recently, dense models have replaced sparse methods as the de facto retrieval method. Rather than focusing on lexical overlap to determine similarity, dense methods build an encoding function that captures semantic similarity by learning from a small collection of question-answer or question-context pairs.

In this paper, we investigate dense retrieval models in the context of open-domain question answering across different input distributions. To do this, first we introduce an entity-rich question answering dataset constructed from Wikidata facts and demonstrate dense models are unable to generalize to unseen input question distributions. Second, we perform analyses aimed at better understanding the source of the problem and propose new training techniques to improve out-of-domain performance on a wide variety of datasets. We encourage the field to further investigate the creation of a single, universal dense retrieval model that generalizes well across all input distributions.

}
\acknowledgements{I would like to thank my wonderful advisor Danqi Chen for giving me the opportunity to work in natural language processing and distilling what it means to be a researcher. I'd would also like to thank my thesis reader Karthik Narasimhan and my fantastic collaborators Zexuan Zhong and Jinhyuk Lee for their insights, knowledge, and discussions. Finally, thank you to all my friends and family for their love and support throughout graduate school.}

\begin{document}

%!TEX root = ../thesis.tex
\chapter{Introduction}

Interest in open-domain question answering has exploded in recent years after the success of DrQA's~\citep{chen-etal-2017-reading} simple neural retriever-reader model. Previous systems were incredibly complex, incorporating tens of subcomponents that would only analyze the question, let alone answer it. Since then, new innovations in the natural language processing community like pre-trained language models have further pushed the field forward, both in terms of efficiency and efficacy.

Early neural models relied on sparse, bag-of-words modeling techniques like TF-IDF and BM25~\citep{robertson2009probabilistic} for efficient document retrieval. While these models are lexically precise and capture general topics well, they struggle to capture semantic meaning. For example, the words ``bad guy'' are encoded completely differently than the word ``villain,'' even though they're semantically identical. To fix this, more recent systems (\citealt{lee-etal-2019-latent}; \citealt{guu2020realm}; \citealt{karpukhin2020dense};  \citealt{xiong2020approximate}) adopt dense retrieval, which incorporates semantic understanding into the encoding step instead of relying on word-word overlap. Dense models leverage pre-trained language models and learn a semantic embedding space in a supervised fashion with labeled question-answer pairs or question-context pairs.

Even though recent approaches post impressive results on common benchmarks, in this paper we demonstrate dense models are far from perfect. The main drawback of dense models is what we denote \textit{out-of-domain (OOD) generalization}, or the model's inability to handle any domain shift from the training distribution. When a model is trained on one dataset and evaluated on an unseen dataset, performance is much worse compared to training and evaluating on the unseen dataset. We show an example in Table \ref{tab:OODBaselineResults} where we compare a dense model trained on one dataset and evaluate on three new datasets with dense models trained on each dataset individually.

%!TEX root = ../thesis.tex

\begin{table*}[!t]
    \centering
    \begin{tabular}{l cc | cc | cc | cc}
        \toprule
        \tf{Models}
        & \multicolumn{2}{c}{NaturalQ} & \multicolumn{2}{c}{TriviaQA} & \multicolumn{2}{c}{WebQ} & \multicolumn{2}{c}{SQuAD} \\
        & R@20 & R@100 & R@20 & R@100 & R@20 & R@100 & R@20 & R@100\\
        \midrule
        DPR (in-domain)$^{\dagger}$ & 80.1 & 86.1 & 79.4 & 85.0 & 73.2 & 81.4 & 63.2 & 77.2 \\
        DPR (pt, NQ) & 80.1 & 86.1 & 70.4 & 79.8 & 68.9 & 78.3 & 48.9 & 65.2 \\
        \midrule
        \midrule
        Difference & 0.0 & 0.0 & \tf{-9.0} & \tf{-5.2} & \tf{-4.3} & \tf{-3.1} & \tf{-14.3} & \tf{-12.0} \\
        \bottomrule
    \end{tabular}
    \caption{Out-of-domain Generalization of DPR. When trained only on one dataset, performance on new datasets is much lower when compared to models trained on the new datasets. We use the released pre-trained (denoted pt) DPR model trained only on NQ released with the code. $^\dagger$: As reported in \citet{karpukhin2020dense}.}
    \label{tab:OODBaselineResults}
\end{table*}

% More concretely, consider a model trained on dataset $D_1$, denoted $M_{D_1}$. Common practice is to segment $D_1$ into 3 sets: training, development, and testing. When evaluated on the test set of $D_1$, $M_{D_1}$ performs spectacularly; however, consider a different dataset $D_2$ and another model trained on $D_2$, $M_{D_2}$. When tested on $D_2$, $M_{D_1}$ underperforms $M_{D_2}$, often by a large margin.

Although this idea has been attributed to numerous issues including unintentional train-test overlap~\citep{lewis2020question} or spurious correlation, lack of generalization prohibits dense models from replacing sparse models in real-world deployment settings. Since sparse models are simply bag-of-words models, they are unable to ``hyperspecialize'' in one particular domain. Using a single model that adapts well across all possible input distributions is an important property for production retrieval systems to have.

Our goal is to help close this gap and push the field towards the creation of a single, universal dense model that is able to perform well on all types of questions. Through a series of experiments and analyses, we aim to better understand both the reasons for this gap and the techniques used to train dense retrieval models.

First, we create a lexically rich QA dataset based on facts from Wikidata that probes how dense models adapt to different relations and to new entities. We empirically observe dense retrieval models perform much worse than sparse models on this benchmark, sometimes by as much as 20\% absolute, even though the questions are very simple. We also observe that simply augmenting a portion of the training dataset helps improve accuracy on seen relations, but does not generalize to unseen relations. Further, we decouple the entity and relation information to demonstrate dense models struggle more when generalizing to unseen relations as opposed to unseen entities.

Second, we perform a variety of experiments on the dense model training process aimed at understanding some of the factors contributing to out-of-domain generalization issues. We explore numerous avenues including removing positional information, fixing encoders during training, using a single encoder instead of a dual encoder, recent innovations in contrastive learning like stop-gradient \citep{chen2020exploring}, novel training datasets, and query-side fine-tuning~\citep{lee2021learning}.

We hope this analysis helps shed light on dense retrieval models and takes one step towards creating a universal dense retriever.

%!TEX root = ../thesis.tex

\chapter{Background}

\section{Open-domain Question Answering}
Open-domain question answering (QA) is a challenging task that takes in a text question $q$ and a large, unstructured text corpus $\mathcal{D}$ and predicts the correct text answer $a$. Formally, we denote the question and answer as sequences of tokens $q = q_1, q_2, ..., q_{l_q}$ and $a = a_1, a_2, ..., a_{l_a}$ where $l_q$ and $l_a$ denote the length of the question and answer sequences respectively. Each document in the text corpus can be segmented into passages, which we denote as the set $\mathcal{P} = \{ p^{(1)}, p^{(2)}, ..., p^{(|\mathcal{P}|)}\}$ where each passage $p^{(i)}$ is a sequence of $l_{p}$ tokens $p^{(i)}_1, p^{(i)}_2, ..., p^{(i)}_{l_p}$. We follow previous work~\citep{karpukhin2020dense} and assume each passage has the same length.

While there are many ways to model this problem, by far the most dominant and widely-used models rely on the \textit{retriever-reader} architecture, first popularized by DrQA~\citep{chen-etal-2017-reading}. Formally, we consider a probability distribution $P(a | q)$ and decompose it into $P(p|q)\cdot P(a|q,p)$. Using this decomposition, the predicted answer to the question would just be:

\begin{equation}
    \underset{a}{\arg\max} \quad P(a|q)=P(p|q)\cdot P(a|q,p),
\end{equation}

where $p$ is a latent variable denoting the evidence passage retrieved as a basis for the answer. In this formulation, systems split the problem into two parts: retrieving relevant evidence $P(p | q)$ and reading the passages $P(a|q, p)$.

The retrieval step considers all passages in the corpus $\mathcal{P}$ and returns a small number of relevant candidates for further processing. The reading step considers the small set of candidates and performs more expensive, but also more expressive, neural reading comprehension to identify the correct answer. Note the the retrieval step acts as a strict upper bound on the overall system performance, as questions with irrelevant retrieved candidates are impossible to answer correctly.

%%%%%%%%%%%%%%%%%%%%%%%%%%%%%%%%%%%%%%%%%%%%%%%%%%%%%%%%%%%%%%%%%
%%%%%%%%%%%%%%%%%%%%%%%%%%%%%%%%%%%%%%%%%%%%%%%%%%%%%%%%%%%%%%%%%

\paragraph{Passage Retrieval}
This paper primarily focuses on the retrieval stage. Formally, we define a retriever $\mathcal{R}$ that takes as input the question $q$ and passage corpus $\mathcal{P}$ and returns a small set of candidates $\mathcal{C}$, where $|\mathcal{C}| \ll |\mathcal{P}|$. We deem a passage \textit{relevant} to the question if the passage contains the answer token sequence.

Retrievers encode the question and each passage into a vector space using an encoding function $f:x\rightarrow\mathbb{R}^H$ where $x$ denotes an arbitrary input token sequence and $H$ denotes the encoded dimension. For a particular question, we approximate the probability distribution $P(p|q) \propto S(q, p)$ where $S(q, p)$ denotes the inner product between the encoded question and the encoded passage. Specifically:

\begin{equation}
    S(q, p) = f_q(q) \cdot f_p(p),
\end{equation}

where $f_q$ and $f_p$ are potentially distinct encoding functions for the query and passage respectively. The retriever collects the candidate set by selecting the top-$|\mathcal{C}|$ highest-scoring passage in the knowledge source. For efficient computation, the passages are typically preprocessed and indexed offline, while the question encoding and search takes place online.

%%%%%%%%%%%%%%%%%%%%%%%%%%%%%%%%%%%%%%%%%%%%%%%%%%%%%%%%%%%%%%%%%
%%%%%%%%%%%%%%%%%%%%%%%%%%%%%%%%%%%%%%%%%%%%%%%%%%%%%%%%%%%%%%%%%

\paragraph{Evaluation Metric}
In general, the goal of a good retriever is to maximize the number of input questions where at least one returned passage is relevant. In our results, we consider recall-at-$K$, denoted $R@K$, which evaluates the percentage of examples that retrieve at least one passage with the correct answer within the first $K$ candidate results. We optimize this metric in our retrieval step in order to maximize the number of examples the reader model can answer correctly.

%%%%%%%%%%%%%%%%%%%%%%%%%%%%%%%%%%%%%%%%%%%%%%%%%%%%%%%%%%%%%%%%%
%%%%%%%%%%%%%%%%%%%%%%%%%%%%%%%%%%%%%%%%%%%%%%%%%%%%%%%%%%%%%%%%%

\section{Sparse Retrieval} Many retrieval encoding functions are based on sparse bag-of-words representations. Formally, we define a sparse encoding function as $f_{sp}:x\rightarrow\mathbb{R}^{|V|}$ where $x$ denotes an arbitrary input token sequence and $|V|$ is the size of the unigram vocabulary $V$. In our formulation, we consider a unigram vocabulary, however equivalent formulations exist for larger vocabularies such as bigrams or arbitrary $n$-grams.

An example of a sparse encoding scheme is TF-IDF, which can be decomposed into a dot product between two values: a term frequency vector \textsc{tf}, and an inverse document frequency vector \textsc{idf} (hence the name). The term frequency vector $\textsc{tf} \in \mathbb{R}^{|V|}$ considers each term $t \in V$ and sets the corresponding value for $t$ in the vector to be proportional to the number of times the term occurs in the input sequence. Similarly, the $\textsc{idf}\in\mathbb{R}^{|V|}$ vector considers each term and sets the corresponding value to be inversely proportional to the number of unique passages the term occurs in. The entire encoding scheme can be written as follows:

\begin{equation}
    f_{\textsc{tf-idf}}(x) = \frac{\textsc{tf}(x) \cdot \textsc{idf}(x)}{|\textsc{tf}(x) \cdot \textsc{idf}(x)|},
\end{equation}

where $x$ denotes an arbitrary input token sequence. Using this formulation, the TF-IDF score $S_{\textsc{tf-idf}}$ for a particular query $q$ and passage $p$ would be:

\begin{equation}
    S_{\textsc{tf-idf}}(q, p) = f_{\textsc{tf-idf}}(q) \cdot f_{\textsc{tf-idf}}(p)
\end{equation}

In our experiments, we use \textsc{BM25}, which can be interpreted as the above \textsc{TF-IDF} model with an additional weighting term.

%%%%%%%%%%%%%%%%%%%%%%%%%%%%%%%%%%%%%%%%%%%%%%%%%%%%%%%%%%%%%%%%%
%%%%%%%%%%%%%%%%%%%%%%%%%%%%%%%%%%%%%%%%%%%%%%%%%%%%%%%%%%%%%%%%%

\section{Dense Retrieval}
Most modern retrievers today rely on dense representations to encode queries and documents into a low-dimensional vector space describing semantic meaning. Dense models are built on top of innovations in other areas of the natural language processing community like large pre-trained language models.

\paragraph{BERT} Dense retrievers today use Bidirectional Encoder Representations from Transformers, commonly referred to as  BERT~\citep{devlin-etal-2019-bert}, as the backbone to obtain dense representations. BERT consists of a stack of encoders based on the Transformer architecture~\citep{vaswani-et-al-attention} which use multi-head self-attention to learn powerful representations over the input sequence. 

BERT performs large-scale pre-training using the Masked Language Modeling (MLM) objective. The task replaces $15\%$ of tokens in the input sequence with a special \texttt{[MASK]} token, and the goal is for the model to predict the original token given the surrounding context. BERT also pre-trains on the Next Sentence Prediction (NSP) objective, where the model is given two sentences and needs to determine if one sentence follows the other.

BERT tokenizes input sequences by adding a special \texttt{[CLS]} token to the start of the sequence and a special \texttt{[SEP]} token to the end. BERT encodes the input sequence $x_{\texttt{[CLS]}},x_1, ..., x_n, x_{\texttt{[SEP]}}$ into contextualized hidden representations $h_{\texttt{[CLS]}}, h_1, ..., h_n, h_{\texttt{[SEP]}}$. The \texttt{[CLS]} token encodes a summary representation of the entire sequence while the \texttt{[SEP]} token is used to separate sequences, or denote the end of a sequence.

%%%%%%%%%%%%%%%%%%%%%%%%%%%%%%%%%%%%%%%%%%%%%%%%%%%%%%%%%%%%%%%%%
%%%%%%%%%%%%%%%%%%%%%%%%%%%%%%%%%%%%%%%%%%%%%%%%%%%%%%%%%%%%%%%%%

\paragraph{Dual Encoder Models}
Concretely, a dense encoding function is defined as $f_{de}:x\rightarrow\mathbb{R}^d$ for an input token sequence $x$ and dimension $d$ where $d \ll |V|$. For a token sequence $x_{\texttt{[CLS]}}, x_1, ..., x_n, x_{\texttt{[SEP]}}$, retrievers use BERT to obtain corresponding hidden representations $h_{\texttt{[CLS]}}, h_1, ..., h_n, h_\sep$.

Rather than use all hidden representations to represent the sequence, most retrievers will compress the information using a reduction function \texttt{reduce} that outputs a single hidden vector, usually the representation of the \texttt{[CLS]} token. A general dense encoding function can be implemented as follows:

\begin{equation}
    f_{de}(x) = \texttt{reduce}(\textsc{bert}(x))
\end{equation}
\begin{equation}
    \texttt{reduce}(h_{\cls}, h_1, ..., h_\sep) = h_{\cls},
\end{equation}

where $x$ is an input token sequence. Most models follow a dual encoder architecture, where one BERT model encodes the query and a separate BERT model encodes the passage. For a particular query, we can calculate the dense retrieval score $S_{de}$ for a query $q$ and passage $p$ as follows:

\begin{equation}
    S_{de}(q, p) = f_{de}^{(Q)}(q) \cdot f_{de}^{(P)}(p),
\end{equation}

where the superscript $^{(Q)}$ and $^{(P)}$ denote distinct BERT encoder models. Note that in most dense models, the passage is tokenized as the concatenation of the title of the article it comes from and its contents, separated by an \texttt{[SEP]} token.

% The first model to follow this paradigm in the open-domain QA space is ORQA~\cite{lee-etal-2019-latent}. The authors model the problem using two independent BERT encoders, one for the query and another for the document. The authors also segment Wikipedia into ``evidence blocks'' where each is 288 byte-pair encoded (BPE) tokens. The model incorporates an expensive pre-training step based on the Inverse Cloze Task (ICT), where the goal is to predict the context surrounding a token sequence. After pre-training, ORQA pre-processes all of Wikipedia into a large dense index where the model can perform efficient maximum inner product search (MIPS). Note this means that the document encoder is fixed for the rest of training.

% The model jointly fine-tunes the question encoder and a third BERT encoder, used for neural reading to identify the answer sub-sequence, on the open-domain QA task. The later introduced REALM~\cite{guu2020realm} system adds additional pre-training task and asynchronous index updates to further improve the system. For additional details, we refer the reader to their paper.

\paragraph{Dense Passage Retriever (DPR)} \citet{karpukhin2020dense} use the dual encoder architecture proposed above for dense retrieval and post impressive results in the space of open-domain QA. The authors consider a training dataset where each question has one positive passage $p^+$ and $N$ negative passages $p^-_1, ..., p^-_N$. The loss function optimizes the negative log-likelihood of the positive passage $p^+$, specifically:

\begin{equation}
    \mathcal{L}_{\textsc{dpr}} = -\log \frac{e^{S(q, p^+)}}{e^{S(q, p^+)} + \sum_{i=1}^N e^{S(q, p^-_i)}}
\end{equation}

% They use a novel contrastive learning objective to efficiently train their model using supervised positive question, passage pairs and distantly supervised negative passages.

% Soon afterwards, Dense Passage Retriever (DPR) introduced by \citet{karpukhin2020dense} demonstrated that expensive ICT pre-training step is not necessary to learn retrieval encoder representations. The authors model the system architecture in a similar way, using two independent encoders where one encodes the query and the other encodes the document; however, they train these encoders using a contrastive learning objective instead.

Positive passages come from annotated open-domain QA datasets like Natural Questions (NQ) \citep{kwiatkowski2019natural}, which contain (question, answer, context) triples. Their best performing model considers negative passages from two sources: in-batch negatives and BM25 hard negatives.

In-batch negatives mean that for each question, all of the positive passages for other questions in the same training minibatch are treated as negatives. BM25 hard negatives are high-scoring passages retrieved using BM25 that do not contain the correct answer. In practice, the authors use a batch size of 128 and sample 1 BM25 hard negative per question, leading to $N=255$ effective negatives per question.

DPR segments Wikipedia into 100 token passages and filters out semi-structured data like tables and lists. After training, the model pre-processes a dense document index using FAISS~\citep{faiss2017johnson} for efficient maximum inner product search (MIPS).

After its release, DPR became the primary retriever for many future open-domain QA systems like RAG~\citep{lewis2021retrievalaugmented} and the current state-of-the-art model, fusion-in-decoder (FiD)~\citep{izacard2021leveraging}. There has also been a lot of investigation into mining harder negatives~\citep{xiong2020approximate}, which further improve the performance of DPR. Most of these new models incorporate techniques like generative readers or asynchronously updating the document index, which are orthogonal to our investigation here.

%%%%%%%%%%%%%%%%%%%%%%%%%%%%%%%%%%%%%%%%%%%%%%%%%%%%%%%%%%%%%%%%%
%%%%%%%%%%%%%%%%%%%%%%%%%%%%%%%%%%%%%%%%%%%%%%%%%%%%%%%%%%%%%%%%%

%!TEX root = ../thesis.tex

\chapter{Shortcomings of Dense Retrievers}

%%%%%%%%%%%%%%%%%%%%%%%%%%%%%%%%%%%%%%%%%%%%%%%%%%%%%%%%%%%%%%%%%
%%%%%%%%%%%%%%%%%%%%%%%%%%%%%%%%%%%%%%%%%%%%%%%%%%%%%%%%%%%%%%%%%

\section{Datasets} To evaluate out-of-domain generalization, we consider a wide variety of datasets sourced from different places. A summary of the datasets can be found in Table~\ref{tab:DatasetStatistics}.

\paragraph{Natural Questions (NQ)} \citet{kwiatkowski2019natural} built the Natural Questions dataset using anonymized Google search data. As with previous works, we follow \citet{lee-etal-2019-latent} and use their Natural Questions Open dataset, which filters out questions without short answers and questions with shorts answers longer than 5 tokens.

\paragraph{TriviaQA} \citet{joshi2017triviaqa} introduced TriviaQA, a dataset of trivia questions scraped from the web. We follow previous work and consider only question-answer pairs, discarding their evidence documents.

\paragraph{Web Questions (WQ)} \citep{berant-etal-2013-webq} gather questions from the Google Suggest API into the Web Questions dataset, where answers are entities in Freebase.

\paragraph{CuratedTREC (TREC)} \citet{curatedtrec-2015-baudis} built the CuratedTREC dataset, which is based on the TREC QA tracks. The authors source their queries from numerous online entities like AskJeeves or MSNSearch.

\paragraph{SQuAD} \citet{rajpurkar2016squad} introduced the widely-used SQuAD reading comprehension dataset. It was constructed from crowdsourced workers asking questions about Wikipedia passages presented to them. Following previous work, we consider the SQuAD Open variant, which ignores context passages during evaluation.

%%%%%%%%%%%%%%%%%%%%%%%%%%%%%%%%%%%%%%%%%%%%%%%%%%%%%%%%%%%%%%%%%
%%%%%%%%%%%%%%%%%%%%%%%%%%%%%%%%%%%%%%%%%%%%%%%%%%%%%%%%%%%%%%%%%

\section{T-REx QA Dataset Evaluation}
To test how well models are able to adapt to new settings, we create a QA dataset based on facts from Wikidata~\citep{wikidata2014vrandecic}, a large collection of (subject, relation, object) triples mined from Wikipedia. We sample from the T-REx~\citep{elsahar-etal-2018-rex} dataset, which is a subset of 11M Wikidata triples with aligned sentences.

The full T-REx dataset considers 43 relations; however, we sample 14. We use hand-crafted query templates to rewrite each (subject, relation, object) triple into a question where the subject is part of the question and the object is the answer. Since the relations are very simple (e.g ``Where was [X] born?''), but the subjects are specific entities (e.g. ``Nikolai Arnoldovich Petrov''), we consider this a lexically rich evaluation set.

We further segment the 14 relations into two sets of 7 relations, one that can be seen during training and one that cannot, which we denote as \texttt{seen} and \texttt{unseen} respectively. For the seen relations, we perform an 80/10/10 split for train/dev/test sets, sampled equally from each relation. For the unseen relations, we only construct a test set using 10\% of examples, again sampled evenly from each relation. More details of the query templates, sampled relations, and sizes can be found in Table \ref{tab:TRExDatasetDetails}.

%!TEX root = ../thesis.tex

\begin{table*}[!t]
    \centering
    \resizebox{0.95\columnwidth}{!}{
    \begin{tabular}{l | l | l | r | c}
    \toprule
    \tf{Rel.} & \tf{Label} & \tf{Query template} & \tf{Size} & \tf{UN} \\
    \midrule
    P19 & place of birth & Where was [X] born? & 10,000 & \xmark \\
    P159 & headquarters location & Where is the headquarter of [X]? & 10,000 & \xmark\\
    P176 & manufacturer & Which company is [X] produced by? & 10,000 & \xmark\\
    P264 & record label & What music label is [X] represented by? & 10,000 & \xmark\\
    P407 & language of work or name & Which language was [X] written in? & 6,722 & \xmark\\
    P413 & position played on team / speciality & What position does [X] play? & 10,000 & \xmark\\
    P740 & location of formation & Where was [X] founded? & 9,415 & \xmark\\
    \midrule
    P17 & country & Which country is [X] located in? & 10,000 & \cmark \\
    P20 & place of death & Where did [X] die? & 10,000 & \cmark \\
    P30 & continent & Which continent is [X] located? & 10,000 & \cmark \\
    P127 & owned by & Who owns [X]? & 10,000 & \cmark \\
    P136 & genre & What type of music does [X] play? & 10,000 & \cmark \\
    P276 & location & Where is [X] located? & 10,000 & \cmark\\
    P495 & country of origin & Which country was [X] created in? & 10,000 & \cmark \\
    \midrule
    All &  &  & 136,137 \\
    
    \bottomrule
    \end{tabular}
    }
    \caption{T-REx QA Dataset Overview. Relations and query templates used to construct the T-REx QA dataset along with the number of examples per relation. UN denotes whether the relation is included in the unseen test set.}
    \label{tab:TRExDatasetDetails}
\end{table*}

We consider 3 dense models: DPR (pt, NQ) is a pre-trained DPR model trained on only NQ; DPR (pt, Multi) is a pre-trained DPR model trained on NQ, TriviaQA, WebQ, and CuratedTREC in a multi-dataset fashion; and REALM (pt, NQ) is another dual encoder model from \citet{guu2020realm} with intermediate pre-training tasks and joint fine-tuning of the reader and retriever models on NQ. Note that REALM retrieves 288 BPE token ``blocks,'' whereas DPR retrieves 100 word passages, so REALM retrieves more content per passage. As a sparse model, we consider the Pyserini~\citep{lin2021pyserini} implementation of BM25. We adopt all default parameters and we build the index using DPR passage splits.

We also include two additional baselines where we take the DPR (pt, NQ) model and fine-tune it for 10 additional epochs. The $+$ (ft, T-REx) model fine-tunes using only the T-REx training set. The $+$ (ft, NQ+T-REx) model fine-tunes on the union of the NQ and T-REx training sets in a multi-dataset training setup. Details on the fine-tuning setup can be found in Table \ref{tab:TrainingHyperparameters}. We report results in Table \ref{tab:TRExBaselineResults}.

%!TEX root = ../thesis.tex

\begin{table*}[!t]
    \centering
    \resizebox{0.95\columnwidth}{!}{
    \begin{tabular}{l cc cc cc cc}
        \toprule
        & \multicolumn{2}{c}{NaturalQ (NQ)} & \multicolumn{2}{c}{TriviaQA} &  \multicolumn{2}{c}{T-REx (se)} & \multicolumn{2}{c}{T-REx (un)} \\
        & R@5 & R@20 & R@5 & R@20 & R@5 & R@20 & R@5 & R@20 \\
        \midrule
        DPR (pt, NQ)    & 68.3 & \tf{80.1} & 57.0 & 69.0 & 34.2 & 48.2 & 43.9 & 59.0 \\
        DPR (pt, Multi) & 67.1 & 79.5 & \tf{71.3} & \tf{80.0} & 42.9 & 56.4 & 50.3 & 63.6 \\
        REALM (pt, NQ)*  & \tf{70.1} & 79.0 & 69.6 & 77.8 & 41.5 & 54.8 & 57.5 & 70.4 \\
        \midrule
        Init: DPR (pt, NQ) \\
        \midrule
        \quad + (ft, T-REx) & 45.5 & 62.3 & 50.6 & 64.8 & \tf{72.8} & \tf{82.3} & 52.4 & 65.3 \\
        \quad + (ft, NQ+T-REx) & 63.7 & 76.3 & 53.4 & 66.2 & 62.8 & 74.9 & 45.3 & 60.7 \\
        \midrule
        BM25        & 45.3 & 64.5 & 69.4 & 78.6 & 54.4 & 64.4 & \tf{62.7} & \tf{73.8} \\
        \bottomrule
    \end{tabular}
    }
    \caption{Baseline Results on T-REx QA Dataset. \texttt{se} and \texttt{un} denote the seen relation evaluation set and unseen relation evaluation set respectively. *: REALM considers 288 BPE token blocks whereas DPR and our BM25 index use 100 word passages.}
    \label{tab:TRExBaselineResults}
\end{table*}

While dense models perform well on NQ and TriviaQA, they significantly underperform on the T-REx QA subsets. It's also notable that REALM \textit{still} underperforms BM25, even though it retrieves more tokens per passage and incorporates expensive intermediate pre-training regimes. This demonstrates that dense models miss key information that sparse models are able to pick up in order to answer these questions.

Looking at the fine-tuned baselines, augmenting examples from T-REx improves performance on the seen relation subset enormously, even outperforming the sparse model; however, if we only fine-tune using T-REx, accuracy on NQ and TriviaQA degrades heavily. When fine-tuning on both NQ and T-REx, we avoid the degradation on NQ and TriviaQA with most, but not all, of the improvements on the seen relation subset. In both cases, very little performance gains on the seen relation subset translate to the unseen relation subset, which means the knowledge learned does not transfer to new relations. These results indicate that current data augmentation techniques or multi-dataset training setups are not enough to close the out-of-domain generalization gap.

%%%%%%%%%%%%%%%%%%%%%%%%%%%%%%%%%%%%%%%%%%%%%%%%%%%%%%%%%%%%%%%%%
%%%%%%%%%%%%%%%%%%%%%%%%%%%%%%%%%%%%%%%%%%%%%%%%%%%%%%%%%%%%%%%%%

\section{Entities vs. Relations}
The questions in the T-REx QA dataset have two distinct dimensions: the subject entities referenced and the specific relations tested. We aim to decouple these two aspects in order to see whether dense models struggle to generalize on unseen relations or on unseen entities.

We construct 4 different subsets: (seen entities, seen relations), (seen entities, unseen relations), (unseen entities, seen relations), and (unseen entities, unseen relations). For each subset, we consider either the 7 seen relations or the 7 unseen relations and sample 300 QA pairs per relation.\footnote{For the (seen entities, unseen relations) subset, two relations did not have enough overlapping entities, causing this subset to be slightly smaller. Results are still clear and significant.}

We consider 3 models: DPR (rt, NQ) is a re-trained version of DPR trained on NQ that serves as a baseline; DPR (rt, NQ+T-REx) is a re-trained DPR trained on the union of the NQ and T-REx training sets; and BM25. Hyperparameters for the re-trained model variants are included in Table \ref{tab:TrainingHyperparameters}. We present R@5 and R@20 results in Table \ref{tab:EntityRelationAnalysis}. 

%!TEX root = ../thesis.tex

\begin{table*}[!t]
    \centering
    \begin{tabular}{l cc | cc | cc | cc}
        \toprule
        \tf{Model} & \multicolumn{2}{c}{(E: \cmark, R: \cmark)} & \multicolumn{2}{c}{(E: \cmark, R: \xmark)} &  \multicolumn{2}{c}{(E: \xmark, R: \cmark)} & \multicolumn{2}{c}{(E: \xmark, R: \xmark)} \\
        & R@5 & R@20 & R@5 & R@20 & R@5 & R@20 & R@5 & R@20 \\
        \midrule
        DPR (rt, NQ)        & 31.9 & 45.8 & 31.5 & 43.2 & 32.7 & 46.8 & 41.2 & 55.3 \\
        DPR (rt, NQ+T-REx)   & \textbf{69.1} & \textbf{79.5} & 40.1 & 52.6 & \textbf{64.8} & \textbf{75.9} & 44.4 & 60.0 \\
        BM25                & 54.6 & 64.1 & \textbf{48.1} & \textbf{58.5} & 55.9 & 66.1 & \textbf{62.2} & \textbf{73.8} \\
        \bottomrule
    \end{tabular}
    \caption{T-REx Entity/Relation Analysis. In column headers, ``E:'' denotes whether the entities are seen during training and ``R:'' denotes whether the relations are seen during training. \tf{Bold} indicates highest performing model in column.}
    \label{tab:EntityRelationAnalysis}
\end{table*}

It's clear that observing the entities and relations during training significantly improves performance of dense models, even outperforming sparse models. Looking at the (E: \xmark, R: \xmark) column, it's also clear that training on the T-REx training data does not generalize to unseen entities or unseen relations.

When observing entities during training but not relations, accuracy improves meaningfully over the baseline; however, when observing relations during training and not entities, accuracy improves significantly, almost to the levels of observing both relations and entities. This indicates that dense models are able to generalize to unseen entities well using the same relations, but they struggle to generalize on unseen relations, even if these relations include entities seen during training.

%!TEX root = ../thesis.tex

\chapter{Understanding the Problem}

%%%%%%%%%%%%%%%%%%%%%%%%%%%%%%%%%%%%%%%%%%%%%%%%%%%%%%%%%%%%%%%%%
%%%%%%%%%%%%%%%%%%%%%%%%%%%%%%%%%%%%%%%%%%%%%%%%%%%%%%%%%%%%%%%%%

\section{Removing Positional Biases}
One difference between dense and sparse models is the bag-of-words modeling assumption. Sparse models treat all words in the sequence independently and only consider statistics based on term and document frequencies. This completely removes the interactions between words (outside co-occurrence) as well as word compositionality. Dense models, on the other hand, consider the sequence as a whole using BERT and encode word order using positional embeddings.

We investigate whether this bag-of-words modeling assumption, specifically the lack of positional information, helps sparse models generalize better to new distributions. One consideration is that questions from one dataset are written completely differently than questions from a different dataset. Compare the examples in Table~\ref{tab:DatasetSamples}, specifically between TriviaQA and NQ. Questions in TriviaQA are typically very long, robust, and detailed. On the other hand, questions in the NQ dataset are short, fragmented, and occasionally ungrammatical. Training a model that only sees one type of question would likely have trouble generalizing to the other.

To do this, we consider \textit{sequence shuffling}, where we split each sequence by spaces and randomly order the words. Note that this removes the word compositionality and may even break the meaning of the question. We consider shuffling the \textit{question tokens} in the training dataset, denoted as models with \texttt{shuffleQ}. We also consider shuffling the \textit{passage tokens} in the training dataset, denoted as models with \texttt{shuffleP}. All models are based on the re-trained DPR model trained on the NQ dataset, denoted DPR (rt, NQ), and we report R@5 and R@20 on NQ, TriviaQA, WebQ, TREC, and SQuAD in Table  \ref{tab:ShuffleCrossDatasetGeneralization}.

%!TEX root = ../thesis.tex

\begin{table*}[!t]
    \centering
    \begin{tabular}{l | ccccc }
        \toprule
        \tf{Model} & NQ & TriviaQA & WebQ & TREC & SQuAD \\
        \midrule
        $R@5$  \\
        \midrule
        DPR (rt, NQ) &          62.1 & 49.6 & 49.3 & 69.7 & 27.4 \\
        DPR (rt, shuffleQ) &     62.2 & 48.6 & 49.7 & 67.6 & 26.8\\
        DPR (rt, shuffleQ, shuffleP) & 3.6  & 7.9 & 3.8 & 11.4 & 2.1 \\
        \midrule
        $R@20$ \\
        \midrule
        DPR (rt, NQ) &                      75.0 & 63.2 & 63.8 & 82.0 & 43.9 \\
        DPR (rt, NQ, shuffleQ) &            75.2 & 62.6 & 69.3 & 80.8 & 42.5 \\
        DPR (rt, NQ, shuffleQ, shuffleP) &  7.9  & 12.8 & 9.6  & 23.5 & 5.9 \\
        \bottomrule
    \end{tabular}
    \caption{Sequence Shuffling Results. \texttt{shuffleQ} denotes shuffling training questions and \texttt{shuffleP} denotes shuffling training passages.}
    \label{tab:ShuffleCrossDatasetGeneralization}
\end{table*}

Shuffling the question tokens during training doesn't \textit{hurt} accuracy, which means that the model uses very little word composition and essentially ignores positional information altogether. This is notable as word order often changes the meaning or intention of the question, especially around words like ``not'' or when considering multi-word entities.

On the other hand, shuffling the question tokens during training doesn't \textit{help} accuracy, which means the model is not overfitting to the phrasing or formatting of a particular dataset. From these results, the question format differences between NQ and TriviaQA do not affect the model's ability to retrieve relevant information.

Once the positive/negative passage tokens are shuffled during training, performance degrades significantly. This follows intuition since passages are 100 tokens, likely spanning multiple sentences. By breaking the ordering in the passages, most of the meaning will be lost, which is what makes BERT so strong. BERT builds a vector space based on semantics, which is much more difficult to construct without word ordering.

We use these results as a basis to conclude that the positional information in passages during training is very important for BERT to build a semantic vector space; however, positional information in questions is generally unimportant, neither helping nor hurting model generalization.

%%%%%%%%%%%%%%%%%%%%%%%%%%%%%%%%%%%%%%%%%%%%%%%%%%%%%%%%%%%%%%%%%
%%%%%%%%%%%%%%%%%%%%%%%%%%%%%%%%%%%%%%%%%%%%%%%%%%%%%%%%%%%%%%%%%

\section{Freeze One Encoder During Fine-tuning}
We analyze the typical dual encoder architecture to determine what's more important: fine-tuning the question encoder or fine-tuning the passage encoder. To do this, we again consider fine-tuning on top of the pre-trained DPR model trained on NQ, denoted DPR (pt, NQ).

We consider fine-tuning under three conditions: $+$ (ft, T-REx) serves as a baseline and denotes fine-tuning both encoders normally; $+$ (ft, T-REx, fixP) denotes freezing the weights of the passage encoder during fine-tuning, only applying updates to the query encoder; $+$ (ft, T-REx, fixQ) denotes freezing the weights of the query encoder during fine-tuning, only applying updates to the passage encoder. Fine-tuning settings can be found in Table \ref{tab:TrainingHyperparameters}. We report R@5 and R@20 results on NQ, TriviaQA, and both T-REx evaluation subsets in Table \ref{tab:FixEncoderResults}.

%!TEX root = ../thesis.tex

\begin{table*}[!t]
    \centering
    \resizebox{0.95\columnwidth}{!}{
    \begin{tabular}{l cc cc cc cc}
        \toprule
        & \multicolumn{2}{c}{NaturalQ (NQ)} & \multicolumn{2}{c}{TriviaQA} &  \multicolumn{2}{c}{T-REx (se)} & \multicolumn{2}{c}{T-REx (un)} \\
        & R@5 & R@20 & R@5 & R@20 & R@5 & R@20 & R@5 & R@20 \\
        \midrule
        Init: DPR (pt, NQ) & 68.3 & 80.1 & 57.0 & 69.0 & 34.2 & 48.2 & 43.9 & 59.0 \\
        \midrule
        \quad + (ft, T-REx)          & 45.5 & 62.3 & 50.6 & 64.8 & \tf{72.8} & \tf{82.3} & 52.4 & 65.3 \\
        \quad + (ft, T-REx, fixP)  & \tf{60.4} & \tf{75.0} & \tf{53.7} & \tf{66.6} & 50.1 & 63.5 & 46.5 & 59.9 \\
        \quad + (ft, T-REx, fixQ)    & 51.9 & 68.3 & 51.8 & 65.4 & 71.5 & 81.5 & \tf{53.8} & \tf{67.5} \\
        \bottomrule
    \end{tabular}
    }
    \caption{Freeze One Encoder During Fine-tuning Results. All models initialized from pre-trained DPR trained on NQ only. \texttt{fixQ} denotes fixing weights in the query encoder and \texttt{fixP} denotes fixing weights in the passage encoder. \texttt{se} denotes seen relation evaluation set and \texttt{un} denotes unseen relation evaluation set. \tf{Bold} indicates highest accuracy model in column.}
    \label{tab:FixEncoderResults}
\end{table*}

We notice that there is a discrepancy between training both encoders, training only the passage encoder, and training only the query encoder. When fine-tuning on T-REx, freezing the passage encoder and training only the query encoder improves performance meaningfully on the T-REx seen relation subset while only degrading slightly on NQ. When freezing the query encoder and only training the passage encoder, accuracy on the T-REx subsets matches that of training both encoders. Interestingly, NQ performance does not degrade as significantly on this model compared to training both encoders, even though T-REx performance is almost identical. We also note improvement in the unseen relation subset compared to training both encoders. Based on these results, we conclude that the context encoder is particularly important to better answer questions from the T-REx QA dataset.

%!TEX root = ../thesis.tex

\chapter{Approaches}

%%%%%%%%%%%%%%%%%%%%%%%%%%%%%%%%%%%%%%%%%%%%%%%%%%%%%%%%%%%%%%%%%

\section{Modified Training Techniques}
We consider DPR and modify the proposed training regime to further investigate how the training objective affects generalization. All training hyperparameters can be found in Table \ref{tab:TrainingHyperparameters}.

\paragraph{Single Model Training} We modify DPR's dual encoder architecture by tying the weights of the query encoder and the passage encoder, effectively creating a single model architecture. The core idea here is a single model that encodes both queries and passages can mimic a ``query-aware'' passage encoder and a ``passage-aware'' query encoder, whereas the dual encoder architecture considers each independently. We compare results between the dual encoder architecture and the single encoder architecture. Models using this technique are denoted with \texttt{1enc}.

%%%%%%%%%%%%%%%%%%%%%%%%%%%%%%%%%%%%%%%%%%%%%%%%%%%%%%%%%%%%%%%%%

\paragraph{Stop-gradient Training} Inspired by \citet{chen2020exploring}, we investigate using a new loss function based on the idea of stop-gradient training. Specifically, we define a new loss function:

\begin{equation}
    \mathcal{L}_{\textsc{stopg}} = \frac{1}{2}(q_d \cdot p) + \frac{1}{2}(q \cdot p_d),
\end{equation}

where $q$ denotes the query representation, $p$ denotes the passage representation, and $_d$ denotes detachment from the computational graph. We compare this loss with the unmodified contrastive loss. Models using this technique are denoted with \texttt{stopG}.

%%%%%%%%%%%%%%%%%%%%%%%%%%%%%%%%%%%%%%%%%%%%%%%%%%%%%%%%%%%%%%%%%

\paragraph{PAQ Training} Many open-domain QA systems base their models on the Natural Questions dataset, which sources queries from anonymized Google search data; however, this distribution bakes in its own set of biases on the task. For example, many Google searches ask questions about topics common in pop culture like notable movies, trending celebrities, and famous musicians. Oftentimes, systems trained on this dataset favor certain information over others. We consider training a model using the Probably Asked Questions (PAQ) \citep{lewis2021paq} dataset, a large-scale collection of (question, answer, passage) triples built using a question generation model. 

First, the authors train a classifier to identify passages likely to be asked about based on the NQ dataset. The authors identify probable answer spans in the 10 million most-likely passages using named entity recognition (NER) tools and a learned answer span model trained on NQ. Next, the authors train a query generation model on NQ, TriviaQA, and SQuAD in a multi-dataset fashion. The query-generation model takes (answer, passage) pairs as input and outputs questions about the passage with the corresponding answer. Finally, the authors filter questions by ensuring the state-of-the-art open-domain QA model, fusion-in-decoder (FiD) trained on NQ is able to generate the correct answer given the question, without the associated passage.

For our study, we group all of the questions asked about a particular passage and filter out any passages that have less than 3 generated questions. We then sample 100K such passages and sample one question asked about each. We split this dataset into 70K/15K/15K for train/dev/test splits, although we do not evaluate on this dataset.

We hypothesize that the PAQ dataset has some important benefits for generalization compared to normal open-domain QA datasets. First, the passage distribution is based on 10M passages, which is about half of Wikipedia, as opposed to popular or trendy topics in Natural Questions. Second, the answer distribution considers both named-entity recognition and an answer span model trained on NQ, which is much more robust than just considering one or the other. Third, we argue that the PAQ dataset is similar to a multi-task learning setup because the query generation model is trained on multiple datasets. This allows us to simulate multi-dataset training while still training on a single, reasonably-sized dataset. We investigate the difference between models trained on Natural Questions and models trained on PAQ. Models trained on PAQ are denoted with \texttt{PAQ}.

%%%%%%%%%%%%%%%%%%%%%%%%%%%%%%%%%%%%%%%%%%%%%%%%%%%%%%%%%%%%%%%%%

\paragraph{Flipped Training} We modify the original training objective to consider positive and negatives \textit{questions} for a given \textit{passage}. While the original training objective likely encourages a successful question discriminator, we hope to encourage a good passage discriminator to improve the passage vector space. We use the PAQ training dataset, where we consider the 70K passages and use randomly sampled questions generated for that passage as positives. For negatives, we use randomly sampled questions from other passages in the training set. Note that we do not incorporate hard negative mining for this training dataset. Models using the flipped training objective are denoted with \texttt{flip}.

%%%%%%%%%%%%%%%%%%%%%%%%%%%%%%%%%%%%%%%%%%%%%%%%%%%%%%%%%%%%%%%%%

\paragraph{Results} We re-train models according to the parameters noted in Table \ref{tab:TrainingHyperparameters} and report combined results in Table \ref{tab:NoQsftRetrainedModels}. Using a single encoder during training improves performance across the board, with most of the gains on out-of-domain datasets. We hypothesize this stems from two reasons. First, a query-aware passage encoder is better able to encode relevant information than the dual encoder architecture. Second, using a single encoder helps align the passage vector space and query vector space compared to a dual encoder by using a single model instead of trying to align two distinct models.

%!TEX root = ../thesis.tex

\begin{table*}[!t]
    \centering
    \begin{tabular}{l cc | cc | cc | cc}
        \toprule
        \tf{Models}
        & \multicolumn{2}{c}{NaturalQ (NQ)} & \multicolumn{2}{c}{TriviaQA} & \multicolumn{2}{c}{T-REx (se)} & \multicolumn{2}{c}{T-REx (un)}  \\
        & R@5 & R@20 & R@5 & R@20 & R@5 & R@20 & R@5 & R@20\\
        \midrule
        \tf{Re-trained Dense Models} \\
        \midrule
        DPR (rt, NQ)        & 62.1 & 75.0 & 49.6 & 63.2 & 31.1 & 45.0 & 40.9 & 55.0 \\%
        DPR (rt, NQ, 1enc)  & \underline{\tf{65.9}} & 77.8 & \underline{57.5} & 69.4 & \underline{34.0} & \underline{47.2} & \underline{43.2} & \underline{58.8} \\%
        DPR (rt, NQ, stopG) & 65.4 & 77.4 & 54.3 & 66.6 & 30.3 & 44.1 & 41.2 & 57.0 \\
        DPR (rt, NQ, 1enc, stopG) & \underline{\tf{65.9}} & \underline{\tf{78.0}} & 57.2 & \underline{69.5} & 32.7 & 46.4 & 39.8 & 56.3 \\
        \midrule
        DPR (rt, PAQ)       & 47.8 & 66.8 & 57.2 & 70.4 & 42.4 & 56.8 & 49.0 & 63.0  \\
        DPR (rt, PAQ, flip) & 41.6 & 62.6 & 51.2 & 65.2 & 39.7 & 53.4 & 44.9 & 60.2  \\
        DPR (rt, PAQ, 1enc) & \underline{50.1} & 68.7 & 59.8 & \underline{72.8} & 44.8 & 57.9 & 53.2 & 67.9  \\
        DPR (rt, PAQ, stopG) & 47.5 & 67.6 & 56.0 & 69.9 & 44.8 & 58.6 & 48.5 & 63.7 \\
        DPR (rt, PAQ, 1enc, stopG) & 50.0 & \underline{69.9} & \underline{60.9} & 72.6 & \underline{50.4} & \underline{63.1} & \underline{55.7} & \underline{69.0}  \\
        \midrule
        \midrule
        BM25                & 45.3 & 64.5 & \tf{69.4} & \tf{78.6} & \tf{54.4} & \tf{64.4} & \tf{62.7} & \tf{73.8} \\%
        \bottomrule
    \end{tabular}
    \caption{Re-trained Models with Modifications. \texttt{1enc} denotes single encoder training. \texttt{stopG} denotes loss function inspired by stop-gradient. \texttt{flip} denotes using positive/negative questions for a passage instead of positive/negative passages for a question. \tf{Bold} indicates highest performing model in column. \underline{Underline} denotes highest performing model with same training data.}
    \label{tab:NoQsftRetrainedModels}
\end{table*}

Using the loss function inspired by stop-gradient has very mixed results. On one hand, the dual encoder architecture when trained on NQ improves performance on NQ and TriviaQA, with slight variation on the T-REx QA dataset. The models trained on PAQ also have mixed results, generally showing marginal differences compared to the dual encoder baseline. On the other hand, when combined with the single encoder architecture, results improve slightly, but meaningfully, when trained on PAQ, but do not improve when trained on NQ. In general, we conclude this loss function has minimal effects compared to single encoder training.

Models trained on the PAQ dataset have some interesting characteristics. These models generally perform much worse (typically by 9-10\% absolute) on the NQ dataset, even though the query generation model is trained partly on NQ. PAQ models also perform much better on TriviaQA and T-REx QA compared to NQ models. These characteristics may be influenced by the answer distribution, which may be more entity-heavy than normal datasets due to the NER answer extraction step.

Finally, the flipped training objective underperforms normal DPR training, not matching performance on any of the datasets for either training setting. This could be due to the choice of negatives since \citet{karpukhin2020dense} show that harder negatives improve retrieval performance when compared to randomly sampled negatives.

%%%%%%%%%%%%%%%%%%%%%%%%%%%%%%%%%%%%%%%%%%%%%%%%%%%%%%%%%%%%%%%%%
%%%%%%%%%%%%%%%%%%%%%%%%%%%%%%%%%%%%%%%%%%%%%%%%%%%%%%%%%%%%%%%%%

\section{Query-side Fine-tuning}
First introduced in \citet{lee2021learning}, query-side fine-tuning fixes the passage encoder and only trains the query encoder based on the true retrieval objective in open-domain QA. For each example, the query model encodes the question and performs the full maximum inner product search over the document index to retrieve the top candidate results. The loss $\mathcal{L}_{\textsc{qsft}}$ reinforces positive passages (those that contain the answer) and can be defined as:

\begin{equation}
    \mathcal{L}_{\textsc{qsft}} = -\log \sum_{\substack{c \in \mathcal{C}\\ \textsc{pos}(q, c)}} \frac{\exp(S(q, c))}{\sum_{c \in \mathcal{C}} \exp(S(q, c))},
\end{equation}

where $\mathcal{C}$ denotes the top document candidates, $\textsc{pos}(q, c)$ denotes whether the candidate $c$ has the answer to the question, and $S(q, c)$ denotes the MIPS retrieval score. Query-side fine-tuning helps close the gap between training and testing tasks by performing full retrieval over the document index instead of using static positives and negatives. We consider query-side fine-tuning on all models presented previously using hyperparameters included in Table \ref{tab:TrainingHyperparameters}. We report R@5 and R@20 results in Table \ref{tab:QsftRetrainedModels}.

%!TEX root = ../thesis.tex

\begin{table*}[!t]
    \centering
    \resizebox{0.95\columnwidth}{!}{
    \begin{tabular}{l cc | cc | cc | cc}
        \toprule
        \tf{Models}
        & \multicolumn{2}{c}{NaturalQ} & \multicolumn{2}{c}{TriviaQA} & \multicolumn{2}{c}{T-REx (se)} & \multicolumn{2}{c}{T-REx (un)}  \\
        & R@5 & R@20 & R@5 & R@20 & R@5 & R@20 & R@5 & R@20\\
        \midrule
        \tf{Re-trained Dense Models} \\
        \midrule
        DPR (rt, NQ) + qsft        & 67.3 & 78.9 & 59.4 & 71.3 & \underline{39.2} & \underline{53.6} & 54.8 & 67.4 \\%
        DPR (rt, NQ, 1enc) + qsft  & 68.3 & 79.0 & 60.2 & \underline{72.1} & 37.7 & 52.0 & 54.9 & 67.1 \\%
        DPR (rt, NQ, stopG) + qsft & \underline{\textbf{68.5}} & 78.9 & \underline{60.4} & \underline{72.1} & 37.9 & 51.5 & 53.7 & 67.2 \\
        DPR (rt, NQ, 1enc, stopG) + qsft & 68.0 & \underline{\textbf{79.4}} & \underline{60.4} & 72.0 & 38.6 & 53.0 & \underline{55.3} & \underline{68.1} \\
        \midrule
        DPR (rt, PAQ) + qsft       & \underline{63.9} & \underline{76.7} & \underline{62.3} & \underline{73.8} & \underline{46.1} & 59.9 & 56.3 & 69.2 \\
        DPR (rt, PAQ, flip) + qsft & 64.1 & 76.6 & 61.7 & 73.5 & 44.3 & 58.9 & 53.9 & 68.3  \\
        DPR (rt, PAQ, 1enc) + qsft & 61.1 & 74.9 & 61.2 & 73.1 & 43.2 & 58.1 & 54.0 & 68.2  \\
        DPR (rt, PAQ, stopG) + qsft & 63.4 & 76.5 & 61.7 & 73.2 & 44.8 & 60.0 & 56.9 & 70.4 \\
        DPR (rt, PAQ, 1enc, stopG) + qsft & 62.6 & 75.5 & 61.0 & 73.1 & 44.9 & \underline{61.3} & \underline{59.6} & \underline{71.7}  \\
        \midrule
        \midrule
        BM25                & 45.3 & 64.5 & \tf{69.4} & \tf{78.6} & \tf{54.4} & \tf{64.4} & \tf{62.7} & \tf{73.8} \\%
        \bottomrule
    \end{tabular}
    }
    \caption{Query-side Fine-tuning Results. \texttt{1enc} denotes single encoder training. \texttt{stopG} denotes loss function inspired by stop-gradient. \texttt{flip} denotes using positive/negative questions for a passage instead of positive/negative passages for a question. \texttt{qsft} denotes query-side fine-tuning on the Natural Questions dataset. \tf{Bold} indicates highest performing model in column. \underline{Underline} denotes highest performing model with same training data.}
    \label{tab:QsftRetrainedModels}
\end{table*}

Query-side fine-tuning improves accuracy across all datasets for all models considered. Surprisingly, improvements are especially pronounced on dual encoder architectures, closing the gap in performance with their single encoder counterparts. We also note that the flipped training objective, which underperformed the regular DPR model, now matches performance. 

We hypothesize this improvement across the board is due to two reasons. First, query-side fine-tuning performs retrieval over the passage index instead of on static positives/negatives. This is closer to the actual task performed at inference time, as noted in \citet{lee2021learning}. Second, query-side fine-tuning helps align the vector space of the query encoder with the vector space of the passage encoder. Since the training signal is only propagating through the query encoder, models are able to better shift the vector space to match representations in the fixed dense passage index.

% We note that the DPR (rt, NQ, 1enc, stopG) + qsft model nearly matches the results of the pre-trained NQ model, but improves performance on the T-REx dataset, even though it uses a significantly smaller batch size (24 instead of 128) and trains for half the epochs (20 instead of 40). We strongly believe that using a large batch size during training will further improve performance; however, we lack the resources to verify this empirically here.

%!TEX root = ../thesis.tex

\chapter{Conclusion}
We explore out-of-domain generalization of sparse and dense models in the context of open-domain question answering. We demonstrate that dense models struggle to generalize on unseen relations, and are better able to generalize to unseen entities. We show that query encoders do not overfit to query phrasing distributions, but also consider limited word composition. We show that fine-tuning only the context encoder accounts for most of the performance improvement on the out-of-domain T-REx QA dataset. We finally propose simple modifications to the DPR training objective and demonstrate that query-side fine-tuning helps close the gap on out-of-domain datasets for dense models.

\bibliography{custom}
\bibliographystyle{apalike}

%!TEX root = ../thesis.tex

\chapter*{Appendix}

%!TEX root = ../thesis.tex

\begin{table*}[htp]
    \centering
    \resizebox{0.95\columnwidth}{!}{
    \begin{tabular}{l | cccc }
    \toprule
    \tf{Param.} & Pre-trained (pt) & Re-trained (rt) & Fine-tuned (ft) & Query-side Fine-tuned (qsft) \\
    \midrule
    Initialization & BERT-base & BERT-base & Trained Retriever & Trained Retriever  \\ 
    \# Epochs & 40 & 20 & 10 & 2 \\
    Batch Size & 128 & 24 & 24 & 24 \\
    Learning Rate & $2\times 10^{-5}$ & $2\times 10^{-5}$  & $2\times 10^{-5}$ & $2\times 10^{-5}$ \\
    Warmup Steps & 1,237 & 1,237 & 1,237 & 100 \\
    \# GPUs & 8 $\times$ 32G & 4 $\times$11G & 4$\times$11G & 1$\times$11G \\
    \# Docs Retrieved & - & - & - & 100 \\
    \bottomrule
    \end{tabular}
    }
    \caption{Training Hyperparameters for DPR Models.}
    \label{tab:TrainingHyperparameters}
\end{table*}

%!TEX root = ../thesis.tex

\begin{table*}[htp]
    \centering
    \resizebox{0.95\columnwidth}{!}{
    \begin{tabular}{l | l | l }
    \toprule
    \tf{Dataset} & Sampled Question & Sampled Answer \\
    \midrule
    Natural Questions & 1. the little girl who voiced ducky in the & 1. Judith Eva Barsi \\ 
                      & land before time & \\ 
                      & 2. who gets married in the last episode of the office & 2. [ Angela Martin, Dwight Schrute ] \\
    \midrule
    TriviaQA & 1. In mythology, who was forced to dine luxuriously & 1. Damocles\\
             & beneath a sword, suspended by a single hair? &   \\       
             & 2. Caviar is traditionally made from the eggs of & 2. [ True Sturgeon, Accipenser ]\\
             & what type of fish? & \\
    \midrule
    Web Questions & 1. what is julia gillard famous for? & 1. Prime minister \\
                  & 2. what country is heathrow airport in? & 2. London \\
    \midrule
    CuratedTREC & 1. Who invented basketball? & 1. Naismith \\
                & 2. What does CNN stand for? & 2. Cable News Network \\
    \midrule
    SQuAD & 1. What is the simple name given to the light & 1. the photon \\ 
          & quantum today? & \\
          & 2. What was the highest building in Houston & 2. JPMorgan Chase Tower \\ 
          & completed in 1982? &  \\
    \midrule
    T-REx QA & 1. Where was Dave \& Buster's founded? & 1. Dallas \\
          & 2. What position does Evan Hlavacek play? & 2. wide receiver \\
    \midrule
    PAQ & 1. how much money did netflix spend in 2013 on its & \$3 billion \\
        & international expansion & \\
        & 2. what subject did takeo fukuda specialize in & 2. economics \\
    \bottomrule
    \end{tabular}
    }
    \caption{Dataset Samples. We show two question/answer pairs drawn from each dataset considered. Questions with multiple answers (denoted with lists) only require on correct answer to be considered correct.}
    \label{tab:DatasetSamples}
\end{table*}

%!TEX root = ../thesis.tex

\begin{table*}[htp]
    \centering
\begin{center}
 \begin{tabular}{l | cc c c }
 \toprule
 \tf{Dataset} & \multicolumn{2}{c}{Train} & Development & Test \\
\midrule
Natural Questions & 79,168 & 58,880 & 8,757 & 3,610 \\
TriviaQA & 78,785 & 60,413 & 8,837 & 11,313 \\
Web Questions & 3,417 & 2,474 & 361 & 2,032 \\
CuratedTREC & 1,353 & 1,125 & 133 & 694 \\
SQuAD & 78,713 & 70,096 & 8,886 & 10,570 \\
\midrule
T-REx QA & 52,909 & 38,778 & 6,613 & 6,615$^\star$ / 7,000$^\diamond$ \\
PAQ & 70,000 & 70,000 & 15,000 & 15,000 \\ 
\bottomrule
\end{tabular}
\end{center}
    \caption{Dataset Sizes. First 5 rows are from the original DPR paper, which is where we source our datasets and pre-trained in-domain models. The two columns under training denote before and after filtering questions without a positive passage. $^\star$ denotes \texttt{seen} relation evaluation set. $^\diamond$ denotes \texttt{unseen} relation evaluation set.}
    \label{tab:DatasetStatistics}
\end{table*}

\end{document}